\documentclass[letterpaper,10pt,times,mathptm,psfig,conference]{IEEEtran}
\IEEEoverridecommandlockouts
\usepackage{cite}
\usepackage[letterpaper, left=19mm, right=19mm, top=19mm, bottom=19mm]{geometry}
\usepackage{amsmath,amssymb,amsfonts}
\usepackage{algorithmic}
\usepackage[ruled,linesnumbered]{algorithm2e}
\usepackage[normalem]{ulem}
\usepackage{amssymb}
\usepackage{amsmath}
\usepackage{graphicx}
\usepackage{subcaption}
\usepackage{textcomp}
\usepackage[table]{xcolor}
\usepackage{hyperref}
\usepackage{multirow}
\usepackage{booktabs}
\usepackage{array}
\usepackage{tabularx}
\usepackage{xcolor}
\usepackage{comment}
\usepackage{pifont}
\usepackage{epsfig, amsmath, amssymb, wrapfig}
\def\BibTeX{{\rm B\kern-.05em{\sc i\kern-.025em b}\kern-.08em
    T\kern-.1667em\lower.7ex\hbox{E}\kern-.125emX}}

\newcommand{\namet}{{\textsc{USplit-VQA}}\xspace}

\newcommand{\khalil}[1]{}

\author{
Md Khalid Syfullah and Alvi Ataur Khalil\\
Transformative Innovation for Trustworthy AI and Network Security (TITANS) Lab, \\ Computer Science, Southern Illinois University Carbondale, USA\\
\{mdkhalid.syfullah, a.khalil\}@siu.edu
\vspace{-10pt}
}
\begin{document}
\title{\namet: U-Shaped Split Learning for Visual Question Answering with Contribution-Aware Weighted Aggregation}

\maketitle
\thispagestyle{empty}
\pagestyle{empty}

\begin{abstract}
Visual Question Answering (VQA) systems, jointly interpreting images and natural language queries, hold significant promise across many domains, yet the privacy-sensitive nature of user data creates a fundamental barrier. Centralized training requires access to all data, while federated learning requires each client to host the full model. We propose \namet, a U-shaped split learning framework for privacy-preserving VQA in which each client retains the initial layers and the classification head while the server hosts the computationally heavy intermediate layers, keeping raw inputs and labels on the client device. We further introduce Contribution-Aware Weighted Aggregation (CAWA), a gradient-similarity-based client scoring mechanism designed to reduce the influence of malicious updates. Experiments on four VQA datasets (VQA-RAD, SLAKE, PathVQA, and VizWiz) with two backbones show accuracy gains over Federated Learning for the Custom model and reduced accuracy for BiomedCLIP under the evaluated fixed split, alongside client memory reductions of up to 5.8$\times$ and communication reductions of up to 10.8$\times$. With one malicious client, CAWA reduces the attacker's influence by over 98\%, while experiments at higher corruption levels identify its limitations. Reconstruction experiments further show lower inversion quality under the evaluated attacks.

\end{abstract}


\section{Introduction}
\label{sec:introduction}
With the advancement of Large Language Models (LLMs), the use of Visual Question Answering (VQA) has increased significantly, yet Vision Language Models (VLMs) remain among the most computationally expensive models, often deemed ``AI-complete'' or ``AI-hard''~\cite{zhang2024vision}. Their high cost complicates deployment, and prompt-based access through central servers raises concerns for privacy-sensitive data.

In the medical domain, Med-VQA enables automated clinical decision support over radiology scans, pathology slides, and retinal photographs~\cite{lau2018dataset, liu2021slake}. Recent VLMs achieve strong Med-VQA performance~\cite{zhang2025multimodal}, yet rely on centralized training that aggregates patient data at a single server, conflicting with HIPAA, GDPR, and institutional data governance policies~\cite{rieke2020future}.

Distributed learning partially addresses these issues. Federated Learning (FL) keeps raw data local yet forces clients to host the full VLM, imposing prohibitive demands on clinical edge devices~\cite{mcmahan2017communication}, while Standard Split Learning (SL) offloads computation yet exposes labels to the server~\cite{vepakomma2018split}. Bidirectional Contrastive Split Learning (BiCSL)~\cite{sun2024bidirectional} decouples a VQA model into representation and contrastive modules with bidirectional gradient sharing. None of these simultaneously provides the resource distribution, label localization, and robustness targeted here.

We propose \namet, a U-shaped split learning framework for privacy-sensitive VQA. The client hosts encoders and the classification head while the server handles multimodal fusion, keeping raw images, questions, and clinical labels local. Distributing gradient computation across untrusted clients, however, introduces a vulnerability: malicious or low-quality updates can silently corrupt the shared server model, and the server cannot validate them without client data. We address this with Contribution-Aware Weighted Aggregation (CAWA), a gradient-similarity mechanism scoring each client by its contribution to convergence. The Custom model also enables split-point analysis for server-client load balancing. Our contributions are four-fold:

\begin{itemize}
    \item We propose \namet, a U-shaped split learning framework tailored to VQA that keeps raw inputs and labels on the client.
    \item We introduce CAWA, a split-learning aggregation mechanism that reduces the influence of data poisoning and backdoor attacks through adaptive, streak-based client reputation scoring.
   \item We evaluate \namet on four privacy-sensitive VQA datasets (VQA-RAD, SLAKE, PathVQA, VizWiz) with two backbones: BiomedCLIP~\cite{zhang2025multimodal} and a lightweight custom architecture ($\sim$8.5M parameters) for resource-constrained edge devices.
    \item We benchmark against centralized and FL baselines and evaluate model inversion, gradient inversion, data poisoning, split-point selection, and client scalability.
\end{itemize}

We investigate the following Research Questions (RQs):
\begin{itemize}
    \item \textbf{RQ1:} Can \namet match centralized training and FL while preserving data and label privacy?
    \item \textbf{RQ2:} How does \namet compare with FL and BiCSL in accuracy, communication, and client burden?
    \item \textbf{RQ3:} How robust is \namet against model inversion and gradient inversion attacks?
    \item \textbf{RQ4:} How well does CAWA mitigate data poisoning and backdoor attacks across corruption levels?
\end{itemize}

The remainder of this paper is organized as follows. Section II presents preliminaries, Section III reviews related work and research gaps, Section IV describes the proposed framework, models, and datasets, Section V reports our experimental setup and findings, and Section VI concludes.

\section{Background}
\label{sec:background}
In this section, we discuss several preliminary concepts to facilitate an understanding of distributed learning, attacks, and the positioning of our proposed \namet framework.

\subsection{Distributed Learning Paradigms}
\begin{figure}[t]
    \centering
    \includegraphics[width=\columnwidth]{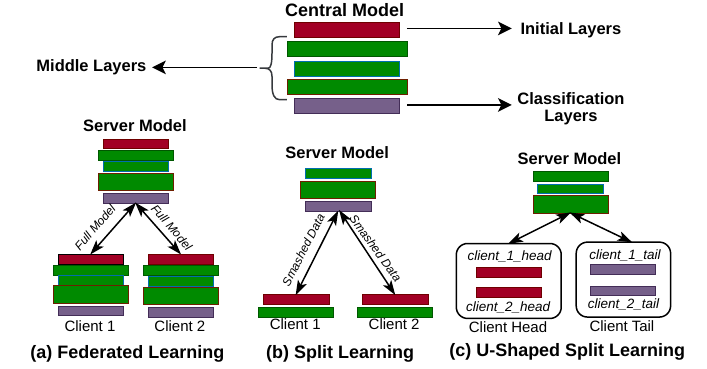}
    \vspace{-15pt}
    \caption{Comparison of architectural layouts.}
    \label{fig:paradigms}
    \vspace{-15pt}
\end{figure}

Distributed learning is a broad AI approach that trains models across multiple computing nodes, typically involving one server and multiple client nodes. Figure~\ref{fig:paradigms} illustrates the three distributed learning paradigms relevant to this work.

\textbf{Federated Learning (FL):} In FL~\cite{mcmahan2017communication}, each client maintains a full copy of the model, trains on private data, and transmits parameter updates to a central server for aggregation. While raw data remains local, clients must host the entire architecture and transmit full model weights.

\textbf{Standard Split Learning (SL):} SL~\cite{vepakomma2018split} partitions the model at a cut layer; the client processes data up to this layer and transmits intermediate activations to the server, which completes the forward pass and computes the loss. This reduces client computation however requires the server to access ground-truth labels, violating privacy.

\textbf{U-Shaped Split Learning:} U-shaped SL~\cite{gupta2018distributed} splits the model twice: the client hosts the bottom encoders and the top classification layers, while the server hosts the middle layers. The forward pass flows client$\rightarrow$server$\rightarrow$client, and loss computation occurs locally. Consequently, neither raw data nor labels are exposed to the server.

\subsection{Threat Models in Distributed Learning}
\label{sec:bg_threats}

Even without explicit data sharing, distributed systems remain vulnerable to various adversarial attacks such as:

\textbf{Gradient Inversion Attacks (GIA):} A passively adversarial server reconstructs private training data from transmitted activations or gradients by optimizing a dummy input to match the observed signals~\cite{zhu2019deep}.

\textbf{Model Inversion Attacks:} An adversary trains a decoder to reconstruct raw inputs from intermediate representations intercepted at the split boundary, exploiting spatial structures retained in early-layer features~\cite{he2019model}.

\textbf{Data Poisoning:} A Byzantine client intentionally corrupts local data or labels (e.g., label flipping or backdoor injection) to submit adversarial gradients, thereby degrading global model performance or installing targeted backdoors~\cite{tolpegin2020data}.


\section{Related Work}
\label{sec:literature-review}
VQA has advanced rapidly through vision-language pretraining: BiomedCLIP~\cite{zhang2025multimodal} improves biomedical representations by pretraining on 15M image-text pairs, and Dong et al.~\cite{dong2025generative} survey recent generative approaches. Existing VQA systems commonly assume centralized data access, leaving a substantial gap in privacy-preserving methodologies.

FL is well studied for unimodal medical imaging~\cite{rieke2020future}, though its application to multimodal Med-VQA remains nascent. Zhu et al.~\cite{zhu2024prompt} proposed prompt-based personalized FL with learnable prompts and a reliability parameter that downweights low-performing clients, and Lu et al.~\cite{lu2023scaling} introduced FedMedVLP to unify disparate medical datasets. Both require clients to host the full VQA architecture, which can be prohibitive for edge devices. SL offers a more resource-efficient alternative and has primarily been explored for unimodal tasks. SplitFed~\cite{thapa2022splitfed} combined split and federated learning for health applications, restricted to single-modality models. For multimodal VQA, Sun et al.\ introduced BiCSL~\cite{sun2024bidirectional}, which relies on contrastive self-supervised learning rather than direct supervised classification, does not employ U-shaped splitting, and is not evaluated on clinical datasets.

Prior work established U-shaped configurations in unimodal distributed settings: RoS-FL~\cite{yang2022robust} partitions U-shaped medical image networks across clients and a server while addressing model drift, UVSL~\cite{wang2024u} protects labels in vertical split learning via local differential privacy, and Lyu et al.~\cite{lyu2023optimal} optimize resource allocation for U-shaped parallel split learning. We tailor this topology to supervised multimodal VQA with contribution-aware aggregation.

The distributed nature of these paradigms also demands robust security measures. Split learning is vulnerable to inversion attacks such as FSHA~\cite{pasquini2021unleashing}, while FL counters data poisoning~\cite{fang2020local} through Byzantine-robust aggregation~\cite{blanchard2017machine}.

Three gaps therefore remain: U-shaped split learning is largely unstudied for supervised multimodal VQA, client contribution scoring in multimodal split learning is underexplored, and a comprehensive cost analysis against FL for VQA is absent. \namet\ addresses these gaps.

\section{Methodology}
\label{sec:methodology}
In this section, we discuss the technical details of our work. Table~\ref{tab:notation} summarizes the key symbols used throughout.

\begin{table}[t]
\centering
\caption{Definitions of used notations and symbols.}
\label{tab:notation}
\footnotesize
\setlength{\tabcolsep}{4pt}
\scriptsize
\begin{tabular}{|c|l|c|l|}
\hline
\textbf{Sym.} & \textbf{Definition} & \textbf{Sym.} & \textbf{Definition} \\
\hline
$K$ & Number of clients & $sim_k$ & Weighted cosine similarity of $k$ \\
\hline
$R, r$ & Total rounds, current round & $c_k$ & Streak counter for client $k$ \\
\hline
$B$ & Batch size & $\tau_{\pm}$ & Adaptive thresholds ($\mu \pm \lambda\sigma$) \\
\hline
$D$ & Hidden dimension (256, 768) & $\rho$ & Temporal factor $(r/R)^\phi$ \\
\hline
$L$ & Text sequence length & $\alpha, \beta$ & Reward / penalty step sizes \\
\hline
$N_v$ & Visual tokens (49, 196) & $\gamma$ & Streak amplification factor \\
\hline
$C$ & Number of answer classes & $\lambda$ & Threshold sensitivity parameter \\
\hline
$V,T$ & Visual / text sequences & $T$ & Softmax temperature \\
\hline
$g_k$ & Server gradient (client $k$) & $\phi$ & Temporal exponent \\
\hline
$R_k$ & Reputation score (client $k$) & $\theta_s, \theta_c$ & Server / client parameters \\
\hline
$W_k$ & Trust weight for client $k$ & $\mathbf{a}_\text{cut}$ & Activation at the cut layer \\
\hline
\end{tabular}
\vspace{-10pt}
\end{table}

\subsection{Model Architectures}
\label{sec:models}
We use a pretrained model (BiomedCLIP~\cite{zhang2025multimodal}) and our defined lightweight custom model, allowing us to observe how \namet performs across different architectures.

\subsubsection{BiomedCLIP-Based Model}
This architecture leverages BiomedCLIP~\cite{zhang2025multimodal} ($D{=}768$), pretrained on PMC-15M. The vision encoder uses ViT-B/16 (196 patch tokens) and the text encoder uses PubMedBERT~\cite{gu2021domain}. Same cross-modal fusion and classification architecture is applied, scaled to $D{=}768$.
\subsubsection{Custom Lightweight Model}
We use a fully trainable architecture ($\sim$8.5M parameters, $D{=}256$) that contains:
 
\noindent \textbf{(i) Vision Encoder:}
A four-stage convolutional pipeline performs aggressive spatial downsampling-
\vspace{-8pt}
\begin{align}
    \mathbf{h}_1 &= \text{MaxPool}\!\left(\phi\!\left(\text{BN}\!\left(\text{Conv}_{7\times7}^{s{=}2}(\mathbf{x})\right)\right)\right) \\
    \mathbf{h}_{l+1} &= \phi\!\left(\text{BN}\!\left(\text{Conv}_{3\times3}^{s{=}2}(\mathbf{h}_l)\right)\right), \quad l = 1, 2, 3 \\
    \mathbf{V} &= \text{LN}\!\left(\text{Reshape}(\mathbf{h}_4)\right) \in \mathbb{R}^{B \times 49 \times D}
\end{align}
\vspace{-2pt}
where $\phi(\cdot)$ is the SiLU activation and the output is a sequence of 49 spatial tokens from a $7{\times}7$ feature grid.
 
\noindent \textbf{(ii) Text Encoder:}
Token embeddings with learned positional encoding are processed through $N_{ce}{=}2$ pre-norm Transformer Blocks (TB):
\vspace{-8pt}
\begin{multline}
    \mathbf{T} = \text{LN}\big(\text{TB}\big(
    \text{LN}(\mathbf{E}_{\text{tok}}[\mathbf{q}] + \mathbf{E}_{\text{pos}}), N_{ce}\big)\big) \in \mathbb{R}^{B \times L \times D}
\end{multline}
 
\noindent \textbf{(iii) CBAM Refinement: }
Visual tokens undergo refinement through $N_{\text{CB}}{=}3$ CBAM blocks~\cite{woo2018cbam}, each applying channel attention $\mathbf{M}_c$ and spatial attention $\mathbf{M}_s$:
\vspace{-5pt}
{
\begin{equation}
    \mathbf{V}^{(l+1)} = \text{LN}\!\left(\mathbf{V}^{(l)} + \text{Flat}(\mathbf{V}_s \odot \mathbf{M}_c \odot \mathbf{M}_s)\right) + \text{FFN}(\cdot)
\end{equation}
}

\noindent \textbf{(iv) Text Refinement:}
An additional $N_T{=}2$ pre-norm transformer blocks deepen text understanding:
\vspace{-5pt}
{
\begin{equation}
    \mathbf{T}^{(l+1)} = \text{LN}\!\left(\mathbf{T}^{(l)} + \text{MHA}(\mathbf{T}^{(l)})\right) + \text{FFN}(\cdot)
\end{equation}
}

\noindent \textbf{(v) Cross-Modal Fusion: }
Four bidirectional cross-attention layers for vision-to-text and text-to-vision fusion:
{\begin{align}
    \mathbf{V}' &= \text{LN}\!\left(\mathbf{V} + \text{MHA}_{V \rightarrow T}(\mathbf{V}, \mathbf{T}, \mathbf{T})\right) + \text{FFN}(\cdot) \\
    \mathbf{T}' &= \text{LN}\!\left(\mathbf{T} + \text{MHA}_{T \rightarrow V}(\mathbf{T}, \mathbf{V}', \mathbf{V}')\right) + \text{FFN}(\cdot)
\end{align}}
A learnable query $\mathbf{q}_p \in \mathbb{R}^{1 \times D}$ extracts a fused representation via cross-attention on concatenated sequence $[\mathbf{V}'; \mathbf{T}']$.
 
\noindent \textbf{(vi) Classification Head: }
The fused vector is mapped to class logits through a two-layer MLP with GELU activation, dropout, and a residual projection.

\begin{figure}[h]
    \centering
    \includegraphics[width=\columnwidth]{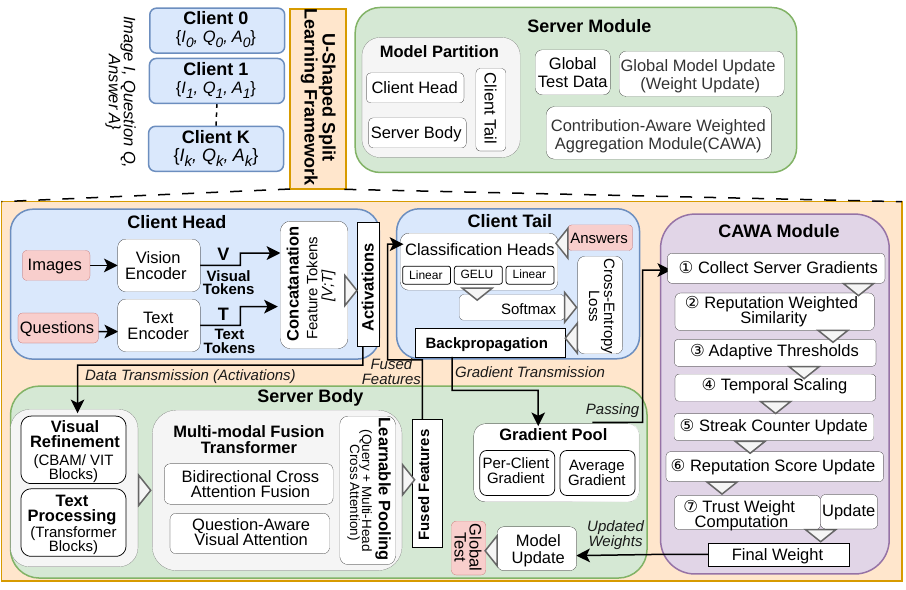}
    \vspace{-15pt}
    \caption{The architectural layout of \namet.}
    \label{fig:ushape}
    \vspace{-15pt}
\end{figure}

\subsection{U-Shaped Split Learning Framework}
\label{sec:usplit}
As illustrated in Figure~\ref{fig:ushape}, the model is partitioned into three components following the U-shaped data flow:

\noindent \textbf{(i) Client Head (forward pass):} The client encodes raw images and questions through local encoders, producing visual tokens $\mathbf{V} \in \mathbb{R}^{B \times N_v \times D}$ and text tokens $\mathbf{T} \in \mathbb{R}^{B \times L \times D}$. These feature tokens are transmitted to the server as activations.

\noindent \textbf{(ii) Server Body (processing):} The server applies visual refinement blocks, text processing blocks, and a multi-modal fusion transformer comprising bidirectional cross-attention, question-aware visual attention, and learnable pooling. The server returns only the fused feature vector $\mathbf{f} \in \mathbb{R}^{B \times D}$.

\noindent \textbf{(iii) Client Tail (backward pass):} The client maps $\mathbf{f}$ through the classification head, computes cross-entropy loss with its private labels, and backpropagates. The logit gradient returns to the server, which updates its parameters through the CAWA module and sends back activation gradients. Raw images, questions, and labels never leave the client. BiomedCLIP uses a fixed split after the first vision and text transformer blocks, placing 38.8M parameters on the client and 187.9M on the server; alternative positions were not evaluated. Training freezes client encoders for 4 rounds, then unfreezes them.

\subsection{Contribution-Aware Weighted Aggregation (CAWA)}
\label{sec:cawa}
In multi-client split learning, uniformly trusting all clients leaves the model vulnerable to adversarial or low-quality updates. CAWA dynamically evaluates gradient alignment against adaptive statistical bounds and scales trust via historical momentum. The protocol is detailed in Algorithm~\ref{alg:usplit_cawa}.
 
\noindent \textbf{(i) Reputation and Loss Weighting: }
Each client maintains an unbounded reputation $R_k$, converted to a bounded trust weight via temperature-scaled softmax:
\vspace{-5pt}
{
\begin{equation}
W_k = \exp\!\left(\frac{R_k - \max_{j} R_j}{T}\right)
\end{equation}
}
This weight directly scales the client's loss: $\mathcal{L}_w = W_k \cdot \mathcal{L}$, suppressing detrimental updates at the gradient level.
 
\noindent \textbf{(ii) Reputation-Weighted Gradient Similarity:}
To prevent a cartel of malicious clients from dominating consensus, the similarity score for client $k$ is the reputation-weighted average cosine similarity with all peers:
\vspace{-5pt}
{\
\begin{equation}
sim_k = \sum_{j \neq k} \tilde{W}_j \frac{\mathbf{g}_k \cdot \mathbf{g}_j}{\|\mathbf{g}_k\| \|\mathbf{g}_j\|}
\end{equation}
}
where $\tilde{W}_j = W_j / \sum_{i \neq k} W_i$ normalizes peer weights.
 
\noindent \textbf{(iii) Adaptive Thresholds and Streak-Based Updates:}
Dynamic bounds $\tau_{\pm} = \mu \pm \lambda\sigma$ (where $\mu, \sigma$ are computed from the current round's similarity distribution) replace static thresholds. A temporal scaling factor $\rho = (r/R)^\phi$ dampens early-round volatility. Reputation is updated using streak-amplified rewards/penalties:
\vspace{-5pt}
{
\begin{equation}
R_k^{(r+1)} = \begin{cases}
    R_k^{(r)} + \rho \alpha e^{\gamma(c_k - 1)} & \text{if } sim_k > \tau_+ \\
    R_k^{(r)} - \rho \beta e^{\gamma(|c_k| - 1)} & \text{if } sim_k < \tau_- \\
    R_k^{(r)} & \text{otherwise}
\end{cases}
\end{equation}
}
where streak counter $c_k$ tracks consecutive positive/negative evaluations. Default hyperparameters are: $\alpha{=}0.1$, $\beta{=}0.05$, $\gamma{=}0.5$, $\lambda{=}0.5$, $T{=}1.0$, and $\phi{=}2.0$.

\begin{algorithm}[t]
\caption{\namet with CAWA}
\label{alg:usplit_cawa}
\begin{algorithmic}[1]
\footnotesize
\renewcommand{\algorithmicrequire}{\textbf{Input:}}
\REQUIRE Clients $K$, Rounds $R$, $\alpha, \beta, \gamma, \lambda, T, \phi$
\STATE Init: $R_k = 0$, $c_k = 0, \forall k$
\FOR{$r = 1, \dots, R$}
    \STATE $R_{\max} \leftarrow \max_j R_j$
    \FOR{each client $k \in K$}
        \STATE $W_k \leftarrow \exp((R_k - R_{\max})/T)$
        \FOR{batch $(\mathbf{x}, \mathbf{q}, \mathbf{m}, \mathbf{y})$ in $k$'s data}
            \STATE $\hat{\mathbf{y}} \leftarrow \text{USplitFwd}(\mathbf{x}, \mathbf{q}, \mathbf{m})$
            \STATE $\mathcal{L}_w \leftarrow W_k \cdot \text{CE}(\mathbf{y}, \hat{\mathbf{y}})$; update $\theta_s, \theta_c$
        \ENDFOR
        \STATE $\mathbf{g}_k \leftarrow$ avg server gradient over batches
    \ENDFOR
    \FOR{each client $k \in K$}
        \STATE $sim_k \leftarrow \sum_{j \neq k} \tilde{W}_j \cos(\mathbf{g}_k, \mathbf{g}_j)$
    \ENDFOR
    \STATE $\mu, \sigma \leftarrow \text{mean}, \text{std}$ of $\{sim_k\}$; $\rho \leftarrow (r/R)^\phi$
    \FOR{each client $k \in K$}
        \IF{$sim_k > \mu + \lambda\sigma$}
            \STATE $c_k \leftarrow \max(1, c_k+1)$; $R_k \mathrel{+}= \rho\alpha e^{\gamma(c_k-1)}$
        \ELSIF{$sim_k < \mu - \lambda\sigma$}
            \STATE $c_k \leftarrow \min(-1, c_k-1)$; $R_k \mathrel{-}= \rho\beta e^{\gamma(|c_k|-1)}$
        \ELSE
            \STATE $c_k \leftarrow 0$
        \ENDIF
    \ENDFOR
\ENDFOR
\end{algorithmic}
\end{algorithm}

\subsection{Datasets Utilized}
\label{sec:datasets}
We evaluate our approach on VQA-RAD~\cite{lau2018dataset}, SLAKE~\cite{liu2021slake}, PathVQA~\cite{he2020pathvqa}, and VizWiz~\cite{gurari2018vizwiz}, as summarized in Table~\ref{tab:datasets}. We use the English SLAKE subset, apply majority voting to VizWiz annotations, and normalize answers into classification categories. Distributed training uses IID client partitions, while test sets are maintained centrally.

 
\begin{table}[h]
\centering
\caption{Summary of the used datasets.}
\label{tab:datasets}
\begin{tabular}{lccccl}
\toprule
\textbf{Dataset} & \textbf{Images} & \textbf{Train} & \textbf{Test} & \textbf{Class} & \textbf{Domain} \\
\midrule
VQA-RAD & 315 & 3,064 & 451 & 490 & Radiology \\
SLAKE & 642 & 4,923 & 1,055 & 218 & Medical \\
PathVQA & 4,998 & 19,755 & 6,279 & 4033 & Pathology \\
VizWiz & 4,319 & 3,455 & 864 & 175 & Accessibility \\
\bottomrule
\end{tabular}
\vspace{-5pt}
\end{table}

\subsection{Split Point \& Client Scalability Analysis}
\label{sec:method_split}
 
The split boundary position determines the privacy-efficiency trade-off. For the Custom model, we evaluate four configurations (V1-V4) that progressively shift model capacity from the server to the client, as summarized in Table~\ref{tab:split_config}. For measuring client scalability, the number of clients was varied from 5 to 15. All preserve the same architecture; only the partitioning differs.

\begin{table}[t]
\centering
\caption{Details for varying model split points.}
\label{tab:split_config}
\setlength{\tabcolsep}{3pt}
\begin{tabular}{clccl}
\toprule
 \textbf{Version} & \textbf{Client Components} & \textbf{Cl.\%} & \textbf{Sv.\%} & \textbf{Transmitted} \\
\midrule
V1 & Vision Enc. & 2.0 & 98.0 & Visual tokens \\
V2 & V1 + Text Embed & 42.6 & 57.4 & Vis. + text tokens \\
V3 & V2 + CBAM + Text Ref. & 63.4 & 36.6 & Refined tokens \\
V4 & V3 + CrossAttn + Fusion & 81.4 & 18.6 & Fused tokens \\
\bottomrule
\end{tabular}
\end{table}


\section{Experimental Results}
\label{sec:experimental_results}
In this section, we present our experimental setup, results, and the strengths and weaknesses of our contribution.

\subsection{Experimental Setup}
\label{sec:setup}

All experiments use PyTorch, with the Custom model on a Tesla T4 GPU and BiomedCLIP on an NVIDIA RTX PRO 6000. We benchmark \namet against centralized training (full data access) and Federated Learning (3 local epochs per round), using $K{=}5$ clients with IID partitioning over 20 rounds, extended to 30 rounds for the scalability study. Hyperparameters: server LR $3{\times}10^{-4}$, client LR $3{\times}10^{-5}$, dropout 0.15, weight decay $10^{-4}$, batch size 24 (Custom) and 16 (BiomedCLIP); centralized models train up to 30 epochs with early stopping (patience 16). Security and split analyses (Sections~\ref{sec:poisoning}--\ref{sec:results_split_scale}) use the Custom model on SLAKE.

\subsection{Accuracy Comparison with Centralized and FL Baselines}
\label{sec:perf}

\begin{table}[t]
\centering
\caption{Comparison of test accuracy (\%) across different datasets, model architectures, and training paradigms.}
\label{tab:main}
\setlength{\tabcolsep}{3pt}
\begin{tabular}{l|ccc|ccc}
\toprule
 & \multicolumn{3}{c|}{\textbf{BiomedCLIP}} & \multicolumn{3}{c}{\textbf{Custom}} \\
\textbf{Dataset} & Cent. & FL & \namet & Cent. & FL & \namet \\
\midrule
VQA-RAD  & 43.68 & 44.57 & 37.92 & 39.25 & 36.36 & 43.90 \\
SLAKE    & 81.24 & 83.51 & 74.18 & 78.23 & 65.88 & 67.77 \\
PathVQA  & 68.42 & 69.18 & 58.25 & 46.88 & 46.60 & 48.25 \\
VizWiz   & 60.89 & 62.62 & 61.01 & 62.25 & 60.02 & 62.75 \\
\bottomrule
\end{tabular}
\vspace{-5pt}
\end{table}

Table~\ref{tab:main} shows that with the Custom model, \namet exceeds FL on all four datasets and matches or exceeds centralized performance on three, with the largest FL gain of 7.54 percentage points on VQA-RAD. Under the fixed BiomedCLIP split, FL performs better on all datasets, with the largest gap of 10.93 points on PathVQA and the smallest of 1.61 on VizWiz. The accuracy impact therefore depends on the backbone and split boundary.

\begin{figure}[ht]
\centering
\includegraphics[width=\columnwidth]{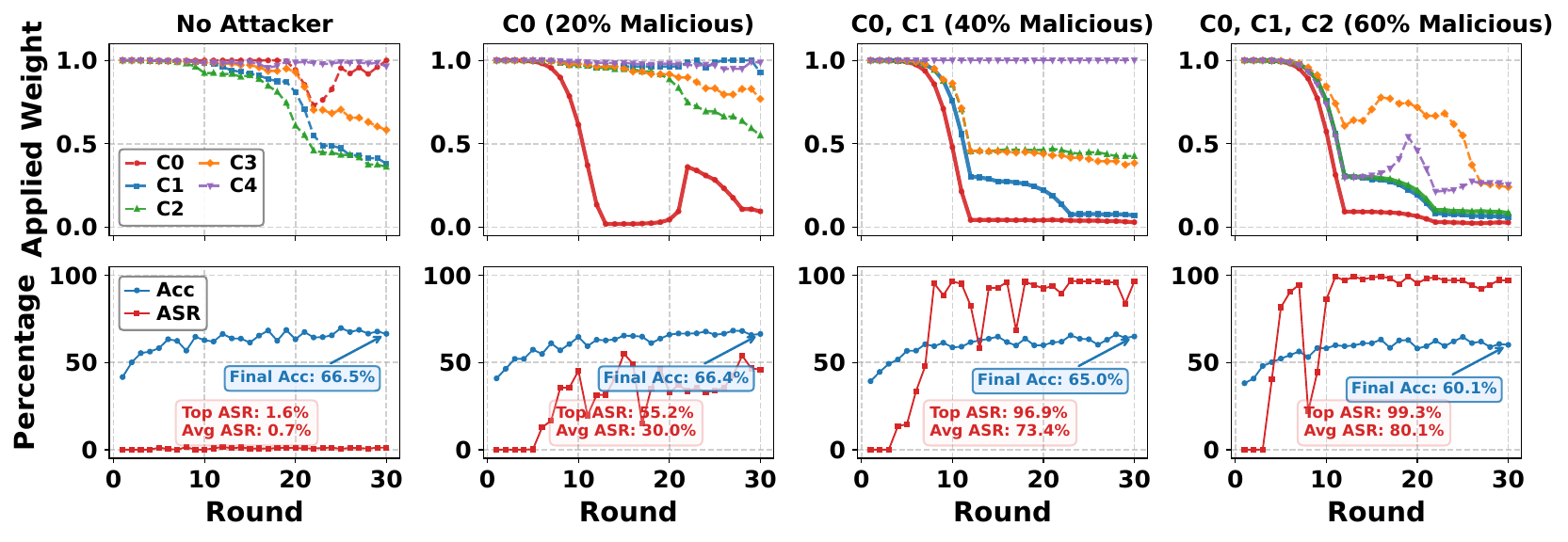}
\caption{Evolution of client reputation weights and overall model performance under various data poisoning intensities.}
\label{fig:cawa}
\end{figure}

\subsection{Comparison with Federated Learning}
\label{sec:fl_comp}

\begin{table}[t]
\centering
\caption{Resource and performance comparison between FL and \namet.}
\label{tab:fl}
\setlength{\tabcolsep}{3pt}
\begin{tabular}{c|cc|cc}
\toprule
 & \multicolumn{2}{c|}{\textbf{BiomedCLIP}} & \multicolumn{2}{c}{\textbf{Custom}} \\
\textbf{Metric} & FL & \namet & FL & \namet \\
\midrule
Client params       & 226.7M    & 38.8M   & 8.5M    & 1.5M    \\
Server params       & n/a       & 187.9M  & n/a     & 7.0M    \\
Client memory       & 2,720\,MB & 465\,MB & 102\,MB & 18\,MB  \\
Comm./round         & 907\,MB   & 84\,MB  & 34\,MB  & 5.6\,MB \\
Time/round (s)      & 139.6     & 99.5    & 68.1    & 36.3    \\
Avg.\ accuracy (\%) & 64.97     & 57.84   & 52.22   & 55.67   \\
\bottomrule
\end{tabular}
\vspace{-10pt}
\end{table}

\begin{figure}[ht]
\centering
\includegraphics[width=\columnwidth]{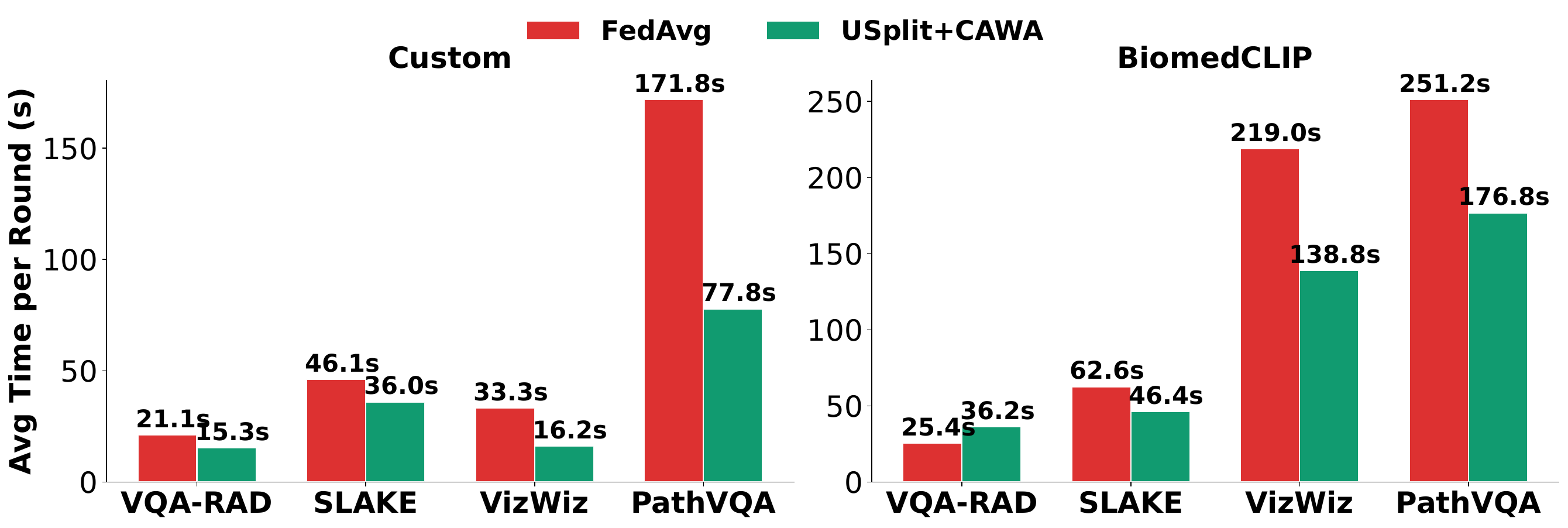}
\caption{Comparison of time requirements per round.}
\label{fig:fl_comp}
\vspace{-10pt}
\end{figure}

Table~\ref{tab:fl} shows that \namet cuts client memory by $5.8\times$ (BiomedCLIP) and $5.7\times$ (Custom), and communication by $10.8\times$ and $6.1\times$ respectively. Average per-round time falls by 47\% for the Custom model and 29\% for BiomedCLIP, although Figure~\ref{fig:fl_comp} shows a BiomedCLIP exception on VQA-RAD, where \namet takes 36.2\,s against 25.4\,s for FedAvg. We estimate client memory as $P_\text{client} \times 4\,\text{B} \times 3$ and communication from full parameter exchange for FL or cut-layer activations and gradients for \namet. \namet keeps labels local, a property FL shares and conventional SL lacks, and CAWA adds Byzantine robustness.

\subsection{Byzantine Robustness Evaluation}
\label{sec:poisoning}

We evaluate Byzantine robustness against a compound attack in which malicious clients flip 30\% of labels and apply an $8{\times}8$ white trigger to 20\% of samples ($y_t{=}1$). Figure~\ref{fig:cawa} shows that CAWA suppresses one attacker's weight by 98.2\% while maintaining 65.2\% accuracy, although performance degrades as the malicious-client proportion rises.

Table~\ref{tab:poisoning} compares \namet, \namet with uniform aggregation (No CAWA), and FL. With one malicious client, \namet records 65.2\% accuracy and 0.6\% ASR against 61.2\% and 44.1\% for the ablated variant. With two or more malicious clients, all frameworks exhibit high ASR, while \namet retains the highest clean accuracy under three malicious clients.

\begin{table}[ht]
\centering
\caption{Clean accuracy and Attack Success Rate (ASR) under varying numbers of malicious clients.}
\label{tab:poisoning}
\setlength{\tabcolsep}{4pt}
\renewcommand{\arraystretch}{1.0}
\begin{tabular}{cl|cccc}
\toprule
\textbf{Method} & \textbf{Metric} & \textbf{No Att.} & \textbf{1 Mal.} & \textbf{2 Mal.} & \textbf{3 Mal.} \\
\midrule
\multirow{2}{*}{\textbf{\namet}}
  & Acc. & 65.7 & 65.2 & 62.6 & 60.9 \\
  & ASR  & 0.6  & 0.6  & 83.5 & 94.2 \\
\cmidrule(lr){1-6}
\multirow{2}{*}{\textbf{No CAWA}}
  & Acc. & 62.5 & 61.2 & 60.6 & 59.2 \\
  & ASR  & 0.6  & 44.1 & 95.3 & 96.3 \\
\cmidrule(lr){1-6}
\multirow{2}{*}{\textbf{FL}}
  & Acc. & 64.2 & 62.7 & 61.3 & 58.0 \\
  & ASR  & 0.6  & 3.9  & 93.0 & 97.0 \\
\bottomrule
\end{tabular}
\vspace{-10pt}
\end{table}

\subsection{Robustness Against Privacy Attacks}
\label{sec:inversion}
We evaluate model inversion with a convolutional decoder and apply Deep Leakage from Gradients (DLG)~\cite{zhu2019deep} to 20 test samples. Table~\ref{tab:attacks} shows that \namet produces $10\times$ higher model-inversion MSE, a 9.9\,dB lower PSNR, and lower SSIM~\cite{wang2004image} than centralized training. For DLG, all PSNR values remain below 8\,dB and \namet records the lowest at 5.70\,dB, indicating lower reconstruction quality within this evaluation rather than a general privacy guarantee.

\begin{table}[ht]
\centering
\caption{Summary of model \& gradient inversion attacks.}
\label{tab:attacks}
\setlength{\tabcolsep}{3pt}
\renewcommand{\arraystretch}{1.0}
\begin{tabular}{l|ccc|ccc}
\toprule
\multicolumn{4}{c|}{\textbf{Model Inversion}} & \multicolumn{3}{c}{\textbf{Gradient Inversion}} \\
\textbf{Metric} & \textbf{Cent.} & \textbf{FL} & \textbf{\namet} &
\textbf{Cent.} & \textbf{FL} & \textbf{\namet} \\
\midrule
MSE $\uparrow$    & 0.0018 & 0.0013 & 0.0175   & 0.248 & 0.253 & 0.272 \\
PSNR $\downarrow$ & 27.42 & 28.80 & 17.56 & 6.08 & 6.00 & 5.70 \\
SSIM $\downarrow$ & 0.989  & 0.988  & 0.796 & - & - & - \\
\bottomrule
\end{tabular}
\vspace{-10pt}
\end{table}

\subsection{Comparison with BiCSL}
\label{sec:bicsl}

\begin{figure}[h]
\centering
\includegraphics[width=0.8\columnwidth]{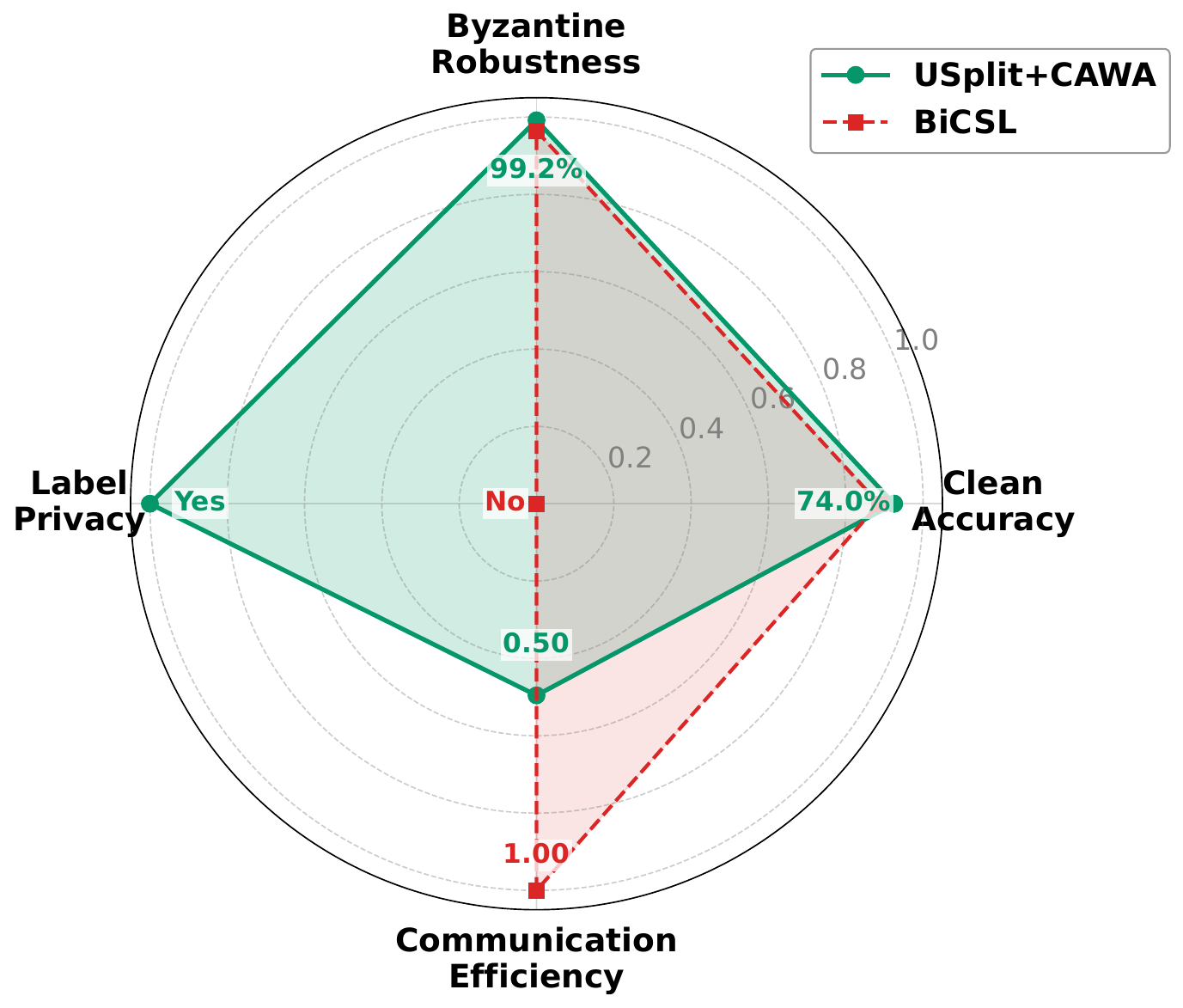}
\vspace{-5pt}
\caption{Trade-offs between \namet and BiCSL.}
\label{fig:radar}
\vspace{-15pt}
\end{figure}

Figure~\ref{fig:radar} compares \namet and BiCSL across four normalized axes. \namet achieves higher clean accuracy and a smaller drop under a 30\% label-flip attack, while keeping ground truth on the client incurs a $2\times$ communication overhead from the U-shaped round trip.

\begin{figure}[h]
\centering
\includegraphics[width=0.95\columnwidth]{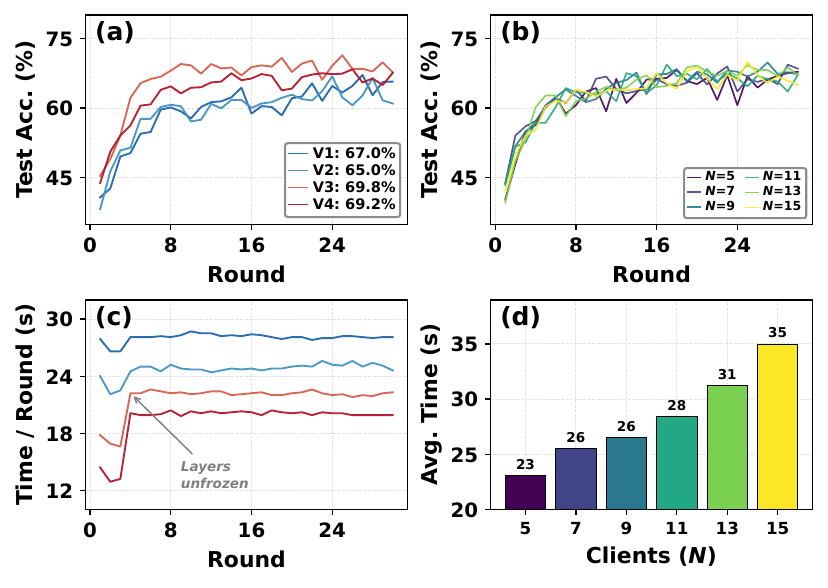}
\vspace{-10pt}
\caption{Split-point (a,\,c) and client-scalability (b,\,d) analysis: test accuracy and per-round time.}
\label{fig:split_client}
\vspace{-15pt}
\end{figure}

\subsection{Split-Point and Client Scalability Analysis}
\label{sec:results_split_scale}

Figure~\ref{fig:split_client} combines the split-configuration study of Table~\ref{tab:split_config} (a,\,c) with the client-scalability analysis (b,\,d). V3 gives the highest accuracy (69.75\%), and V4 is nearly as accurate (69.18\%) with the lowest round time, since a larger client share reduces the activations and gradients exchanged each step. The accuracy and efficiency optima therefore need not coincide, and V3 remains preferable when client resources bind. All $K\in\{5,\dots,15\}$ converge to a 65--70\% accuracy band within 30 rounds (b), while the average per-round time on SLAKE (d) grows sub-linearly from 23.1\,s ($K{=}5$) to 35.0\,s ($K{=}15$), as fixed server-side computation is amortized across clients.

\subsection{Summary of Findings}
\label{sec:discussion}
\namet cuts client memory $5.7$--$5.8\times$ versus FL on both backbones. For \textbf{RQ1} and \textbf{RQ2}, the Custom model exceeds FL in average accuracy, whereas BiomedCLIP falls behind under the evaluated fixed split, indicating a backbone-dependent trade-off. For \textbf{RQ3}, the evaluated inversion attacks reconstruct inputs less faithfully under \namet than under centralized training or FL. For \textbf{RQ4}, CAWA suppresses one malicious client's weight by 98\%, although backdoor protection degrades with two or more attackers.

Several limitations remain. The IID assumption does not reflect clinical heterogeneity, large pretrained backbones lose accuracy under the evaluated U-shaped split, and CAWA weakens as the malicious-client proportion increases. Future work will focus on non-IID adaptation and split-geometry optimization for large-scale transformers.


\section{Conclusion}
\label{sec:conclusion}
We introduced \namet, a U-shaped split learning framework tailored to VQA that keeps raw images, queries, and labels on the client while reducing client memory by up to $5.8\times$ relative to FL. The Custom model improves average accuracy over FL, while BiomedCLIP exhibits an accuracy trade-off under the evaluated fixed split. CAWA strongly suppresses one malicious client, although its backdoor protection declines as the malicious-client proportion increases. The evaluated inversion attacks also produce lower reconstruction quality for \namet. Future research will consider non-IID data and improved split selection for large pretrained models.


\bibliographystyle{IEEEtran}
\bibliography{References}

@article{lau2018dataset,
  title={A dataset of clinically generated visual questions and answers about radiology images},
  author={Lau, Jason J and Gayen, Soumya and Ben Abacha, Asma and Demner-Fushman, Dina},
  journal={Scientific data},
  volume={5},
  number={1},
  pages={180251},
  year={2018},
  publisher={Nature Publishing Group}
}

@inproceedings{liu2021slake,
  title={{SLAKE}: A semantically-labeled knowledge-enhanced dataset for medical visual question answering},
  author={Liu, Bo and Zhan, Li-Ming and Xu, Li and Ma, Lin and Yang, Yan and Wu, Xiao-Ming},
  booktitle={2021 IEEE 18th international symposium on biomedical imaging (ISBI)},
  pages={1650--1654},
  year={2021},
  organization={IEEE}
}

@article{zhang2025multimodal,
  title={A multimodal biomedical foundation model trained from fifteen million image--text pairs},
  author={Zhang, Sheng and Xu, Yanbo and Usuyama, Naoto and Xu, Hanwen and Bagga, Jaspreet and Tinn, Robert and Preston, Sam and Rao, Rajesh and Wei, Mu and Valluri, Naveen and others},
  journal={{NEJM AI}},
  volume={2},
  number={1},
  pages={AIoa2400640},
  year={2025},
  publisher={Massachusetts Medical Society}
}

@article{rieke2020future,
  title={The future of digital health with federated learning},
  author={Rieke, Nicola and Hancox, Jonny and Li, Wenqi and Milletari, Fausto and Roth, Holger R and Albarqouni, Shadi and Bakas, Spyridon and Galtier, Mathieu N and Landman, Bennett A and Maier-Hein, Klaus and others},
  journal={{npj} Digital Medicine},
  volume={3},
  number={1},
  pages={119},
  year={2020},
  publisher={Nature Publishing Group UK London}
}

@inproceedings{mcmahan2017communication,
  title={Communication-efficient learning of deep networks from decentralized data},
  author={McMahan, Brendan and Moore, Eider and Ramage, Daniel and Hampson, Seth and y Arcas, Blaise Aguera},
  booktitle={Artificial intelligence and statistics},
  year={2017},
  organization={{PMLR}}
}

@inproceedings{zhu2024prompt,
  title={Prompt-based personalized federated learning for medical visual question answering},
  author={Zhu, He and Togo, Ren and Ogawa, Takahiro and Haseyama, Miki},
  booktitle={ICASSP 2024-2024 IEEE International Conference on Acoustics, Speech and Signal Processing (ICASSP)},
  pages={1821--1825},
  year={2024},
  organization={IEEE}
}

@article{vepakomma2018split,
  title={Split learning for health: Distributed deep learning without sharing raw patient data},
  author={Vepakomma, Praneeth and Gupta, Otkrist and Swedish, Tristan and Raskar, Ramesh},
  journal={arXiv preprint arXiv:1812.00564},
  year={2018}
}

@article{gupta2018distributed,
  title={Distributed learning of deep neural network over multiple agents},
  author={Gupta, Otkrist and Raskar, Ramesh},
  journal={Journal of Network and Computer Applications},
  volume={116},
  pages={1--8},
  year={2018},
  publisher={Elsevier}
}

@inproceedings{sun2024bidirectional,
  title={Bidirectional contrastive split learning for visual question answering},
  author={Sun, Yuwei and Ochiai, Hideya},
  booktitle={Proceedings of the AAAI Conference on Artificial Intelligence},
  volume={38},
  number={19},
  pages={21602--21609},
  year={2024}
}

@inproceedings{he2019model,
  title={Model inversion attacks against collaborative inference},
  author={He, Zecheng and Zhang, Tianwei and Lee, Ruby B},
  booktitle={Proceedings of the 35th annual computer security applications conference},
  pages={148--162},
  year={2019}
}

@inproceedings{tolpegin2020data,
  title={Data poisoning attacks against federated learning systems},
  author={Tolpegin, Vale and Truex, Stacey and Gursoy, Mehmet Emre and Liu, Ling},
  booktitle={European symposium on research in computer security},
  pages={480--501},
  year={2020},
  organization={Springer}
}

@article{dong2025generative,
  title={Generative models in medical visual question answering: A survey},
  author={Dong, Wenjie and Shen, Shuhao and Han, Yuqiang and Tan, Tao and Wu, Jian and Xu, Hongxia},
  journal={Applied Sciences},
  volume={15},
  number={6},
  pages={2983},
  year={2025},
  publisher={MDPI}
}

@article{lu2023scaling,
  title={Scaling-up medical vision-and-language representation learning with federated learning},
  author={Lu, Siyu and Liu, Zheng and Liu, Tianlin and Zhou, Wangchunshu},
  journal={Engineering Applications of Artificial Intelligence},
  volume={126},
  pages={107037},
  year={2023},
  publisher={Elsevier}
}

@inproceedings{thapa2022splitfed,
  title={{SplitFed}: When Federated Learning Meets Split Learning},
  author={Thapa, Chandra and Arachchige, Pathum Chamikara Mahawaga and Camtepe, Seyit and Sun, Lichao},
  booktitle={Proceedings of the AAAI conference on artificial intelligence},
  volume={36},
  number={8},
  year={2022}
}

@inproceedings{pasquini2021unleashing,
  title={Unleashing the tiger: Inference attacks on split learning},
  author={Pasquini, Dario and Ateniese, Giuseppe and Bernaschi, Massimo},
  booktitle={Proceedings of the 2021 ACM SIGSAC conference on computer and communications security},
  pages={2113--2129},
  year={2021}
}

@inproceedings{fang2020local,
  title={Local model poisoning attacks to {Byzantine-Robust} federated learning},
  author={Fang, Minghong and Cao, Xiaoyu and Jia, Jinyuan and Gong, Neil},
  booktitle={29th USENIX security symposium (USENIX Security 20)},
  pages={1605--1622},
  year={2020}
}

@article{blanchard2017machine,
  title={Machine learning with adversaries: Byzantine tolerant gradient descent},
  author={Blanchard, Peva and El Mhamdi, El Mahdi and Guerraoui, Rachid and Stainer, Julien},
  journal={Advances in neural information processing systems},
  volume={30},
  year={2017}
}

@inproceedings{gurari2018vizwiz,
  title={{VizWiz} Grand Challenge: Answering visual questions from blind people},
  author={Gurari, Danna and Li, Qing and Stangl, Abigale J and Guo, Anhong and Lin, Chi and Grauman, Kristen and Luo, Jiebo and Bigham, Jeffrey P},
  booktitle={Proceedings of the IEEE conference on computer vision and pattern recognition},
  pages={3608--3617},
  year={2018}
}

@inproceedings{woo2018cbam,
  title={{CBAM}: Convolutional Block Attention Module},
  author={Woo, Sanghyun and Park, Jongchan and Lee, Joon-Young and Kweon, In So},
  booktitle={Proceedings of the European conference on computer vision (ECCV)},
  pages={3--19},
  year={2018}
}

@article{gu2021domain,
  title={Domain-specific language model pretraining for biomedical natural language processing},
  author={Gu, Yu and Tinn, Robert and Cheng, Hao and Lucas, Michael and Usuyama, Naoto and Liu, Xiaodong and Naumann, Tristan and Gao, Jianfeng and Poon, Hoifung},
  journal={ACM Transactions on Computing for Healthcare (HEALTH)},
  volume={3},
  number={1},
  pages={1--23},
  year={2021},
  publisher={{ACM}}
}

@inproceedings{zhu2019deep,
  title={Deep Leakage from Gradients},
  author={Zhu, Ligeng and Liu, Zhijian and Han, Song},
  booktitle={Advances in Neural Information Processing Systems},
  volume={32},
  year={2019}
}

@article{wang2004image,
  title={Image quality assessment: from error visibility to structural similarity},
  author={Wang, Zhou and Bovik, Alan C and Sheikh, Hamid R and Simoncelli, Eero P},
  journal={IEEE transactions on image processing},
  volume={13},
  number={4},
  pages={600--612},
  year={2004},
  publisher={IEEE}
}

@article{he2020pathvqa,
  title={{PathVQA}: 30000+ questions for medical visual question answering},
  author={He, Xuehai and Zhang, Yichen and Mou, Luntian and Xing, Eric and Xie, Pengtao},
  journal={arXiv preprint arXiv:2003.10286},
  year={2020}
}

@article{zhang2024vision,
  title={Vision-language models for vision tasks: A survey},
  author={Zhang, Jingyi and Huang, Jiaxing and Jin, Sheng and Lu, Shijian},
  journal={IEEE transactions on pattern analysis and machine intelligence},
  volume={46},
  number={8},
  pages={5625--5644},
  year={2024},
  publisher={IEEE}
}

@article{yang2022robust,
  title={Robust split federated learning for {U}-shaped medical image networks},
  author={Yang, Ziyuan and Chen, Yingyu and Huangfu, Huijie and Ran, Maosong and Wang, Hui and Li, Xiaoxiao and Zhang, Yi},
  journal={arXiv preprint arXiv:2212.06378},
  year={2022}
}

@inproceedings{wang2024u,
  title={{U}-shaped Vertical Split Learning with Local Differential Privacy for Privacy Preserving},
  author={Wang, Liang and Chen, Hao and Zuo, Lina and Liu, Haibo},
  booktitle={International Conference on Intelligent Computing},
  pages={72--81},
  year={2024},
  organization={Springer}
}

@inproceedings{lyu2023optimal,
  title={Optimal resource allocation for {U}-shaped parallel split learning},
  author={Lyu, Song and Lin, Zheng and Qu, Guanqiao and Chen, Xianhao and Huang, Xiaoxia and Li, Pan},
  booktitle={2023 IEEE Globecom Workshops (GC Wkshps)},
  pages={197--202},
  year={2023},
  organization={IEEE}
}

\end{document}